\pdfoutput=1
\documentclass[11pt]{article}
\usepackage{acl}
\usepackage{times}
\usepackage{latexsym}
\usepackage[T1]{fontenc}
\usepackage[utf8]{inputenc}
\usepackage{microtype}
\usepackage{todonotes}
\usepackage{color}
\usepackage{soul}

\newcommand{\copenhagen}{1}
\newcommand{\utrento}{2}
\newcommand{\uppsala}{3}
\newcommand{\leuven}{4}
\newcommand{\sheffield}{5}

\title{Challenges and Strategies in Cross-Cultural NLP}

\author{Daniel Hershcovich$^{\copenhagen}$ ~ Stella Frank$^{\utrento}$ ~ Heather Lent$^{\copenhagen}$ ~ Miryam de Lhoneux$^{\copenhagen,\uppsala,\leuven}$\\ \textbf{Mostafa Abdou$^{\copenhagen}$ ~ Stephanie Brandl$^{\copenhagen}$ ~ Emanuele Bugliarello$^{\copenhagen}$ ~ Laura Cabello Piqueras$^{\copenhagen}$} \\ \textbf{Ilias Chalkidis$^{\copenhagen}$ ~ Ruixiang Cui$^{\copenhagen}$ ~ Constanza Fierro$^{\copenhagen}$ ~ Katerina Margatina$^{\sheffield}$} \\ \textbf{Phillip Rust$^{\copenhagen}$ ~ Anders Søgaard$^{\copenhagen}$}\\
  $^{\copenhagen}$University of Copenhagen \quad
  $^{\utrento}$University of Trento \quad
  $^{\uppsala}$Uppsala University \\
  $^{\leuven}$KU Leuven \quad
  $^{\sheffield}$University of Sheffield \\
  \texttt{dh@di.ku.dk}
 }

\begin{document}
\maketitle
\begin{abstract}
Various efforts in the Natural Language Processing (NLP) community have been made to accommodate linguistic diversity and serve speakers of many different languages. However, it is important to acknowledge that speakers and the content they produce and require, vary not just by language, but also by culture. Although language and culture are tightly linked, there are important differences. Analogous to cross-lingual and multilingual NLP, cross-cultural and multicultural NLP considers these differences in order to better serve users of NLP systems. We propose a principled framework to frame these efforts, and survey existing and potential strategies.
\end{abstract}

\section{A Framework for Cultural Awareness}\label{sec:intro}

Language technology is rapidly advancing for a minority of the world's languages. At the same time, the majority of languages are falling behind \cite{joshi-etal-2020-state}.
It is essential that language technology can serve the speakers of a wide variety of languages, who come from a wide variety of cultures.
In this paper, we argue that doing so requires accommodating these speakers not only on a linguistic level, but also on a cultural level.

Culture, like language, is a term that is hard to pin down, but generally describes the way of life of a collective group of people, and distinguishes them from other groups with other cultures \cite{mora2013cultures,shweder2007cultural}.
Culture encompasses both material as well as non-material aspects, such as beliefs and linguistic practices \cite{kendall2005sociology}.
Moreover, since ``[c]ulture is the acquired knowledge people use to interpret experience and generate behavior'' \cite{spradley72},
it is also the lens through which people understand linguistic messages. 
Culturally maladapted messages can and will be misinterpreted; NLP must be culturally sensitive in order to avoid doing harm.

\begin{figure}[t]
    \centering
    \includegraphics[width=\columnwidth]{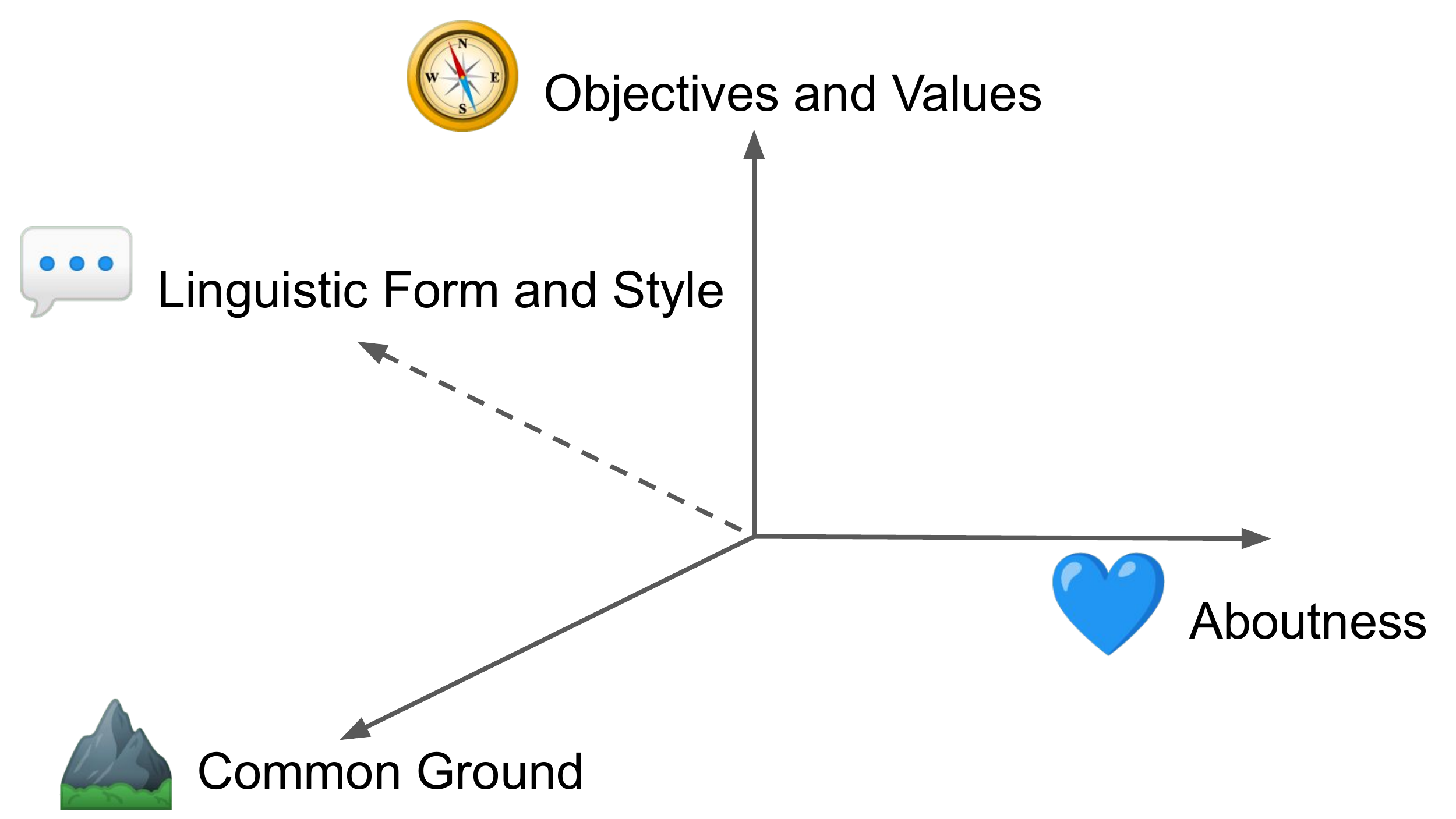}
    \caption{The role of culture in NLP, illustrated by four dimensions along which cultures vary, and for which NLP can be culturally biased: linguistic form and style, common ground, aboutness, and objectives (values).}
    \label{fig:framework}
\end{figure}

Language and culture interact in a number of ways (see Figure~\ref{fig:framework}). This paper aims to illuminate these connections, in order to motivate and inform future NLP work.
Beyond \textbf{linguistic form and style}, \textit{how} things are expressed in language, they include {\bf common ground}, the shared knowledge based on which people reason and communicate; {\bf aboutness}, \textit{what} information people care to convey; and \textbf{objectives} or values, the goals people strive for (e.g., when developing language technology).

Since language and culture are intertwined, the different dimensions may be difficult to tease apart \cite{hovy-yang-2021-importance}:
for example, the lexicon is shaped both by the need to convey information people care about, and by the conceptual categories grounded in the environment.
Style is related to values and societal structure: formality levels in Japanese, for example, reflect the hierarchy of Japanese society \cite{gao2005japanese}. Nevertheless, language and culture are not interchangeable terms. Culture varies greatly \textit{within} languages \cite{lin-etal-2018-mining}.
For example, the assumption that ``English'' in and of itself carries a single set of worldviews, interests and norms is unjustified \cite{paul-girju-2009-cross,wilson-etal-2016-disentangling}.
On the other hand, a relatively homogeneous culture can span multiple languages, as in the Nordic countries \cite{sahlgren-etal-2021-basically}.

\paragraph{Contributions.}
We propose a framework for understanding the challenges that cultural diversity poses for NLP to serve all users, as well as the opportunities that NLP creates to understand these differences better. We consider four elements: linguistic form, common ground, aboutness and values. We then survey existing strategies in NLP to address these challenges. Highlighting limitations in current strategies, as well as successful examples, we propose directions for future development of cross-cultural NLP.

\section{Linguistic Form and Style}\label{sec:style}
Linguistic form refers to non-semantic questions of \emph{how} to formulate an utterance.
Most work on cross-lingual NLP has focused on how linguistic form varies between languages. However, the impact of social and cultural factors on linguistic form and stylistic variations is rarely discussed \cite{hovy-yang-2021-importance}.

\paragraph{Variation within language.}
How to circumscribe and define a particular language is a difficult problem \cite{ethnologue}.
A language spoken in different countries often becomes standardized in slightly different ways (e.g., German in Austria and Germany).
Geo-cultural variation within a language also gives rise to \emph{dialects} 
\cite{zampieri2020,wolfram1997,bookPerspectives}, which in turn operate as an important sign of cultural identity \cite{FALCK2012225}.
In addition, \emph{sociolects} vary across social groups, which can include subcultures \cite{mccormack2011hexagonal}.

Due to these variations, treating ``a language'' as a homogeneous mass limits cultural adaptation, and runs the risk of privileging certain cultures over others.
\citet{zhang-etal-2021-sociolectal} find that pre-trained language models (PLMs; see~\S\ref{sec:strategies}) reflect certain sociolects more than others.
For example, there are considerable morphosyntactic variations between Spanish spoken in Spain and Argentina \citep{bentivoglio2011morphosyntactic}, but they are not considered separately in a Spanish PLM \citep{CaneteCFP2020}. A PLM specific to the Algerian dialect of Arabic performs better than a multilingual or general Arabic one \citep{antoun-etal-2020-arabert} in sentiment classification \citep{abdaoui2021dziribert}.

\paragraph{Stylistic variation.}

Factors such as directness and formality are often associated with different communicative styles across cultures. For example, the level of politeness and thus also how offensive something is perceived depends very much on communicative norms \citep{gao2005japanese,CultureSpecificCo}.
Misunderstandings that arise from different communicative styles can occur in any interaction between people of different cultural backgrounds. \citet{10.1093/applin/4.2.91} refers to this as \textit{pragmatic failure}, namely the ``inability to understand `what is meant by what is said''', due to \emph{how} it is being said.
Comparative stylistics \cite{vinay1995comparative} aims to characterise these differences.

The same intention (e.g., to be polite) can lead to different forms in different cultures.
For example, an offer of help can be made in an imperative form in Polish, while in English speaking Anglo-Saxon cultures, this form would be considered rude and so an interrogative form would be used instead \cite{wierzbicka2009cross}.
Similarly, German native speakers tend to use a high level of directness, which would be considered offensive in English \cite{house2011politeness}.
Based on the observation that such differences exist, \citet{ringel-etal-2019-cross} applied distant supervision by inducing labels for English formality and sarcasm detection based on language---in this case German and Japanese, respectively.

The expression and perception of emotion also varies across cultures, both in text \cite{kirmayer2001cultural, ryder2008cultural} and in face-to-face communication \cite{hareli2015cross}.
These differences are critical for cross-cultural sentiment analysis \cite{bautin2008international} and for text-based recognition of medical conditions such as depression.
For example, \citet{loveys2018cross} find clear differences in linguistic form across English speakers of different ethnic identities with self-reported depression (e.g., the ratio of positive to negative emotion expression).
Not understanding cross-cultural variation could lead to misclassification of movies, or worse, mis-diagnosing people.

\section{Common Ground}\label{sec:ground}

A culture is in part defined by a shared \textit{common ground}: the shared body of knowledge that can be talked about and that can be assumed as known by others.
This common ground varies from culture to culture, and thus cross-cultural language use has to take into account shifts in common ground.

These cross-cultural shifts, often correlating with cross-lingual shifts, are neglected in NLP.
An underlying assumption in many approaches to multilingual NLP is that ``different languages share a similar semantic structure'' \cite{miceli-barone-2016-towards}.
However, the assumption that there is a cross-lingual, cross-culturally common semantics to preserve fails when the common grounding does not match between cultures.
Two relevant aspects here are the set of relevant \textit{concepts}, closely identified with problems of lexicalisation, and \textit{common sense}, i.e., the relevant propositional knowledge used in reasoning and entailment.

\paragraph{Conceptualisation.}

People carve up the world using conceptual categories~\cite{sep-concepts}.
While some general or even universal patterns exist~\cite{wierzbicka1996}, these categories can and do differ between languages and cultures: the domain of colour is a famous example~\cite{berlin69basic},
but cross-cultural differences are also reflected in kinship systems \cite[the lexicalization of family structures][]{murdock70kin}, spatial relations~\cite{bowerman01language} and basic objects~\cite[e.g., where to draw the distinction between a `cup' and a `mug', or the `hand' and the `arm';][]{majid15semantic}.
Translating between languages thus entails translating between conceptualisations, which can be impossible if the conceptual grounding is not available (is this Danish ``kop'' an English `cup' or a `mug?')---this is the motivation behind multimodal, i.e., visually grounded, translation \cite{specia-etal-2016-shared}.
Data collected from a single culture may contain concepts that do not easily map across cultures, as for example seen in the recent MarVL dataset \cite{liu-etal-2021-visually}, in which about a quarter of concepts from a sample of non-Western cultures did not map to English concepts (as represented by WordNet).

\paragraph{Commonsense knowledge.}

``Common sense'' is the knowledge that is held \textit{in common} by a community and culture, the communal knowledge bank that can be presumed to be known by everyone.
Common sense thus covers a diverse set of knowledge types, from physical and temporal reasoning to psychological, social and moral judgements \cite{sap-etal-2020-commonsense}.
Some aspects of common sense are fairly universal, in particular those that arise from inhabiting a grounded human body on earth
(dropping a glass of juice onto a hard surface is likely to cause a mess; the baby might be crying because it's hungry).
On the other hand, axioms of social and moral common sense, for example the how and why of rituals such as marriages, vary between cultures \cite{acharya2021atlas}.

Moving between different banks of common sense can involve different strategies to either provide missing background/common knowledge, or to transfer the content to a target-culture appropriate setting (see~\S\ref{sec:translation}).

\section{Aboutness}\label{sec:about}

Cultures promote different topics and issues \cite{Seibert143}, sometimes by necessity, sometimes by accident.\footnote{What we, as human beings, are interested in is, in part, a product of our group memberships. Our group membership determines conditions, e.g., our dependence on weather, agriculture, livestock, etc., making some topics vital, and promotes others e.g., through fashion, trends, etc. \citet{Heidegger:1927} refers to the former as the {\em facticity} aspect of our {\em caring}; the latter as the {\em fallenness} aspect. Unbiased treatment of the facticity aspect of what cultures care about, has been argued to be particularly important from an ethical perspective \cite{nelson2008heidegger}. }  Some of the domains commonly considered in (English-speaking, Anglo-western) NLP are irrelevant to some cultures, while others {\em mean} different things to different cultures, i.e., they involve completely different practices. An example of the former is beer reviews, a go-to domain in sentiment analysis   \cite{zeng-etal-2019-variational,ji-etal-2020-diversified,paranjape-etal-2020-information}. Beer reviews are hardly meaningful in cultures with no beer consumption.
An example of the latter, i.e., a domain that is used differently across cultures and social groups, is Twitter \cite{doi:10.1080/1461670X.2020.1719369}.  
As an example of multi-domain sentiment analysis, \citet{liu-etal-2018-learning-domain} consider cameras, laptops, restaurants, and movies, within the context of data from U.S. American websites. These domains, however, apply differently in other cultures: restaurants are more important in some cultures (e.g., Copenhagen hipsters) than others (e.g., S{\'a}mi); similarly, laptops have not penetrated cultures equally and have in fact been found to {\em change}~culture \cite{hansen2014how}. 

Beyond conceptualisation differences (see~\S\ref{sec:ground}), \citet{liu-etal-2021-visually} point out how the visual concepts at the core of many multimodal NLP tasks reflect a Northern American and Western European bias in the underlying data sources, from which images were scraped. 
As an alternative, in their new dataset (MarVL), they let the
selection of both concepts and images be entirely
driven by native speakers. This corrects for some existing asymmetries in what our evaluation data tends to be about. For example, \citet{althoff2014howtoaskforafavor} created a dataset entirely devoted to {\em pizzas}; in \citet{liu-etal-2021-visually}, models' ability to recognise {\em vadas}, an Indian food, is evaluated.  

People describe the same events differently across cultures.
News reports emphasise different goals, motivations, methods and content in different parts of the world \citep{Bandura1985SocialFO, Li1997NewVI,Loo2019ReadingV}. However, many news resources used in NLP research reflect only some cultures \citep{Breed1955SocialCI,Galtung1965TheSO, marcus-etal-1993-building}. As NLP technologies have been widely adopted to build multilingual news generators \citep{zhang-etal-2016-towards, xu-etal-2020-xiaomingbot, jin-etal-2020-hooks}, it is important that these differences are taken into account in language generation.

\section{Objectives and Values}\label{sec:values}

Common objectives in NLP include progress, accuracy, fairness, robustness and interpretability.
They are driven by the research community, and reflect the values of this community: novelty, for example, is highly esteemed, which is not a value held by more conservative, tradition-valuing cultures.
Grounding these objectives explicitly in ethical values and norms, and acknowledging that these may differ across cultures, is essential for cross-cultural NLP. \citet{jiang2021delphi}, for example, introduce \textsc{Commonsense Norm Bank} and Delphi, a moral reasoning resource and model, respectively, reflecting ``English-speaking cultures of the United States in the 21st century''. However, researchers, practitioners, regulators and users may belong to different cultures and have different objectives and values \cite{talat+al.local21}.

Another common goal in the NLP community is the eventual expansion of all NLP technologies to low-resource languages. 
However, as \citet{bird-2020-decolonising} points out, this goal must not overshadow, or be placed above, the specific desires and needs of a given language community, as every community will have their own objectives for what they do and do not want for their language with regards to technology. Without building relationships with the communities for which we aim to develop NLP, there is a serious risk for imposing our own objectives on the community.

Importantly, entire categories of applications and tasks may be driven by cultural assumptions. For example, the main objective of text summarisation is brevity \cite{jorgensen-sogaard-2021-evaluation}. Likewise, while fluency is a common objective of machine translation, it may in fact, be less important than comprehensibility for users \cite{Castilho2018}.

\paragraph{Values differ across cultures.}
Values are an important part of non-materialistic culture, as these values define what a specific culture deems to be good or valuable, as opposed to bad and undesirable \cite{kendall2005sociology}. Accordingly, these values help inform what is accepted as normal and ethical behaviour, and shape common cultural attitudes. Aspects that clash against a culture's values, norms, and ethics, may often be taboo or even illegal.
As cultures around the world hold different values, their norms and ethics are inevitably different, too
\cite{doi:10.1177/0956797615586188, 10.3389/fpsyg.2018.00849}. 
For example, in some cultures, alcohol consumption is prohibited for religious reasons, and thus may not be normal or acceptable behaviour; meanwhile drinking alcohol can be seen as normal behaviour by some cultures without this shared value. For this reason, beer reviews in an application, for example, may be seen not only as irrelevant (see \S\ref{sec:about}), but even as offensive.
While these values can and do change greatly from culture to culture, \textit{end-users of NLP systems deserve technologies that are suitable to the culture that they belong to}. 
To this end, the preservation of cultural values, norms, and ethics in models intended for users from different cultures, is another important aspect of cross-cultural NLP \cite{solaiman2021process}.

\paragraph{Fairness and combating biases.}
Many recent studies in NLP have explored strategies to counteract unwanted biases (according to the community's values, see above) in models, resulting from prejudices present in society. These biases are implicit in the data used to train our models, and become baked into the models themselves \cite{NIPS2016_a486cd07,sap-etal-2019-risk,Sun2019-la,blodgett-etal-2020-language,bender2021dangers}. 
Because NLP datasets are composed of utterances from members of a larger culture, the resulting biases constitute a partial reflection of that culture's values, norms, and ethics. 
For example, work in NLP on gender bias demonstrate that there is a pervasive problem with sexism in models (a direct result of sexist attitudes present in society), and explore ways to remedy this \cite{friedman-etal-2019-relating,zhao-etal-2019-gender,50755,baker-gillis-2021-sexism}. 
Accordingly, just as members of any culture deserve language technology that truly serves them, \textit{members of vulnerable groups and marginalised members in larger cultures deserve such as well}.
Current research in addressing fairness and combating biases in models demonstrates how NLP can be used to tackle the challenge of inequitable cultural attitudes. 
In other words, the goal of countering bias in NLP can be seen as shifting existing cultures into hopefully more equitable ones.

\paragraph{Conflicting objectives.}
With regards to cultural values, cross-cultural NLP lies at the intersection of the two important, but potentially conflicting, cross-roads of multicultural pluralism and societal equity: cultures will have different values, which should be respected and represented for end-users; meanwhile, cultures will also have some inequitable attitudes that hurt different end-users, and NLP should be part of the solution.  
Together, these conflicting aspects land cross-cultural NLP in the impossible position of needing to both \textit{preserve} cultural values, while also 
\textit{minimising} harmful cultural biases. 
We acknowledge that these issues become increasingly difficult to disentangle, especially for more difficult or taboo aspects of culture. On the one hand, it is dangerous for members of one culture to impose values onto the others (i.e. this would, in fact, contribute to NLP colonisation---see~\S\ref{sec:conclusion}), and on the other hand, it is dangerous to leave marginalised groups vulnerable, when we can diminish bias against them. 
While none of the strategies we discuss in~\S\ref{sec:strategies} offers a simple solution to this challenge, our goal is to bring this important conundrum in cross-cultural NLP to the community's attention. To help navigate this convoluted space, and thus ensure end-users have access to relevant NLP systems, we encourage people working in this space to collaborate with members of the culture relevant to their work, as they are the best equipped to judge what cultural aspects can appropriately be challenged. 
Additionally, new works in AI ethics should also be considered
\cite{10.1145/3461702.3462608,the_ethics_of_ethics_2020,Institutionalizing_ethics_2021, ethical_issues_across_cultures,2020}.
An international regulatory instrument is essential for the responsible development of AI, a task that UNESCO is in the process of undertaking.\footnote{\href{https://en.unesco.org/courier/2018-3/towards-global-code-ethics-artificial-intelligence-research}{https://en.unesco.org/courier/2018-3/towards-global-code-ethics-artificial-intelligence-research}}

\section{Strategies}\label{sec:strategies}

We have identified several challenges along four dimensions of culture. Here, we highlight general strategies for cross-cultural NLP. We identify three main areas where we could direct efforts towards mitigating cross-cultural disparities: data collection, model training, and translation. Data is the backbone of NLP and any efforts towards cross-cultural NLP needs to consider the strategies involved in collecting and annotating it. Transfer learning is central to cross-lingual NLP and can serve an important role in cross-cultural NLP. Finally, translation is used to communicate between languages and will often be necessary when communicating between cultures. We explain what can be done within each of these areas each in turn.

\subsection{Data Collection}\label{sec:data}

The most fundamental issue is the representation disparity in our data, i.e.,~not all cultures are (equally) represented. While the volume of multilingual datasets increases and multilingual NLU benchmarks are becoming available \cite{hu2020xtreme,liang-etal-2020-xglue,ruder2021}, this does not guarantee cross-cultural representation. There are two main factors: source of data (e.g.,~media outlet) and origin of annotations (e.g.,~automatic vs human-generated). To have truly diverse datasets, we should both ensure that they not only represent diverse sources, but also multiple perspectives in terms of annotations, when applicable.

\paragraph{Data selection and curation.}
Relevant to the impact of the source of data, \citet{dodge2021documentingc4} discuss how the Colossal Clean Crawled Corpus \citep[C4; ][]{2020t5}, an English web-based corpus, is skewed in favour of US governmental institutions and main-stream US media, while data filtering, which tries to remove slurs, or obscene words, disproportionately removes text from and about minorities (e.g., African American and LGBTQ+).
Similarly, data from Wikipedia is heavily used in multilingual NLP even though it has been shown to be culturally biased \citep{Callahan2011CulturalBI}.
To mitigate the risks associated with using culturally biased data, data selection or curation methods should strive to use data sources that are appropriate for the target culture of downstream NLP applications. Large, general-purpose datasets should be curated so as to be as unbiased as possible, and carefully documented \cite{10.1145/3351095.3372829,bender2021dangers}. 

As examples of culturally diverse data collection, \citet{liu-etal-2021-visually} and \citet{yin-etal-2021-broaden} recruit geo-diverse annotators from Africa, Asia and the Americas who are tasked with also providing the data (the images) along with the annotations (image descriptions).
However, these dataset are small and thus limited to evaluation only; the cost and effort associated with collecting sufficient data to train modern NLP models means that finding culture-specific data to cover the enormous diversity of cultures represented on earth remains a formidable challenge.
A diverse and open community, however, facilitates scalability in this regard. Examples for such communities include Universal Dependencies, dedicated to manual annotation of morphosyntactic datasets for over one hundred languages, covering the highest typological diversity to date among such datasets \citep{nivre-etal-2020-universal}. As another example, the Masakhane community aims to strengthen NLP research for African languages. It has created MasakhaNER \citep{masakhaner21}, a Named Entity Recognition dataset for 10 African languages, collected by native speakers of these languages.

\paragraph{Data annotation.}
With respect to annotation practices, a diverse pool of annotators reduces the risk of cultural bias. 
In the case of subjective tasks such as detecting affect, aggression, and hate speech, annotators may systematically disagree with one another due to cultural differences that are often reflected by their biases and values \cite{davani2021dealing}. Annotator disagreements may capture important nuances in such tasks that are often ignored while aggregating annotations to a single (possibly hegemonic) ground truth. Release of all annotations with each dataset, even disagreeing ones, allows training models that generalise better \cite{plank-etal-2014-learning,pavlick-kwiatkowski-2019-inherent,prabhakaran-etal-2021-releasing,gordon2022jury}. Careful documentation of the annotation process also plays an important role. In a recent survey of machine learning papers (including NLP ones), \newcite{geiger2020garbage} reported a wide divergence in the level of documentation about methodological practices in human annotation. They advocate for the importance of human annotation within the research process, arguing that it ought to be given as much attention, care, and concern as is currently placed on performance based metrics.

When arguing for an increasingly diverse `participatory design' \cite{bodker2009participatory}, however, it is important to consider the values, ideologies, codes, narratives, and power relations which govern the interaction between the assemblies of actors involved in data collection and annotation. This is the perspective of social-cultural studies and other critical viewpoints from the humanities \cite{mainsah2014participatory, gray2019ghost}.

\paragraph{Annotation projection.}

Of course, data collection for many languages and cultures can be costly.
Using parallel data and a word alignment tool, annotation (for example a syntactic tree) in a source language can be transferred to a target language without extra annotation \cite{yarowsky-etal-2001-inducing,Hwa2005BootstrappingPV}.
A related method to create multilingual datasets is translating a dataset (typically in English) into other languages (often using machine translation). While not usually called annotation projection, it can be considered a variation of this method, since the source annotation is ported to the translated data. 
For example, in XNLI \citep{conneau-etal-2018-xnli}, the premise and hypothesis pair is translated from English to other languages and the label is reused for the translated pair.
These methods help leverage data from high-resource languages to create more data for low-resource languages \cite{agic-2017-cross}.

However, these methods risk ignoring target culture complexity or forcing the source culture  concepts on the target culture.
For example, English common sense datasets  \citep{singh-etal-2021-com2sense} include culture-specific concepts such as food ingredients,\footnote{\textit{A hot sauce is going to be hotter if it uses jalapeño peppers rather than habanero}.} rituals and celebrations,\footnote{\textit{Colt doesn't have any kids. Finley has four kids. Therefore, Finley is more likely to go Trick or Treating}.} and societal expectations.\footnote{\textit{Many people disapproved of the widow waiting one week after his [sic] wife's death to start dating again, rather than one year}.}
Translating this dataset into other languages will require making decisions about how, and whether, to modify these items to make them more intelligible in the target culture.
\citet{lin-etal-2021-common} use machine translation to translate two common-sense reasoning datasets from English into 14 other languages.
They attempt to deal with difficult cases by automatically flagging and removing examples which contain `social keywords' from the dataset, or that are (again, automatically) labeled as containing non-neutral sentiment.
However, these methods are unlikely to
capture all examples of social behaviour and
cannot identify examples of cultural over-specificity (sports teams, jalapeños).
Automatically translated \textit{training} data can lead to worse performance than native target language data \cite{liu-etal-2021-visually}.
However, if \textit{evaluation} data is automatically translated too, we have no trivial way of exposing cultural biases introduced by the projection process.
Culturally-aware evaluation thus necessitates data annotated directly in the target language, or at least culturally-sensitive human translations.

\citet{ponti-etal-2020-xcopa} point out that literal translation of datasets is sometimes impossible or undesirable due to culture-specific concepts in the source that may be missing or unnatural in the target.
In their multilingual extension of the English Choice of Plausible Alternatives \citep[COPA; ][]{roemmele-2011-copa} dataset, they therefore asked
``carefully chosen'' human translators to perform
culturally-sensitive translation, and either paraphrase, substitute the original concepts with similar ones that exist in the target language, or leverage phonetically transcribed loan words.

Human translation, or original data from the target culture, is clearly the expensive option, but will often be the only way to avoid cultural bias.
Only translating/generating high-quality \textit{evaluation} data is becoming an attractive middle ground option \cite{liu-etal-2021-visually,ponti-etal-2020-xcopa,yin-etal-2021-broaden} that at least allows us to judge the success of cross-lingual transfer in a culturally appropriate way.

\subsection{Model Training}\label{sec:model}

Models can be culturally biased even when trained on culturally diverse data. On the other end, diverse cultural representation can in some cases be achieved with specific training strategies, not only data selection and annotation.

\paragraph{Transfer learning and pre-training.}
A common strategy in machine learning is to \emph{transfer} the knowledge acquired for a task, domain or language to another. It is used extensively in the context of cross-lingual learning, motivated by similarities between languages and cultures: if only the form differs between languages, then models can learn to abstract away from it and transfer from resource-rich languages, obviating task-specific training data in many languages \cite{agic-2017-cross,wu-dredze-2019-beto,blloshmi-etal-2020-xl}.

In this context, approaches for creating cross-lingual word embeddings \citep{klementiev-etal-2012-inducing,Ammar2016MassivelyMW} are based on the assumption that the semantic spaces of different languages are approximately \emph{isomorphic}. However, this assumption is violated in practice \cite{sogaard-etal-2018-limitations,sogaard2019cross}. Among linguistic reasons, cultural factors (e.g., conceptualisation) can play a role in the mismatch between the spaces.
Though hardly ever considered when selecting the source language for model transfer, considering cultural factors improves performance on target languages in pragmatically motivated tasks \cite{sun-etal-2021-cross}.

In recent years, \emph{model transfer} has been the mainstream paradigm with the advent of pre-trained language models (PLMs), and specifically, multilingual PLMs such as mBERT \citep{devlin-etal-2019-bert} and XLM-R \citep{conneau-etal-2020-unsupervised}. Language models require massive amounts of data to \emph{pre-train}.
Since corpora are largely skewed in favour of few languages, this leads to cross-lingual disparities. To mitigate this issue, \citet{NEURIPS2019_c04c19c2} introduced an exponential smoothing of the language \emph{sampling} rate leading to a less skewed data selection. \citet{conneau-etal-2020-unsupervised} and \citet{xue-etal-2021-mt5} further studied the effect of the language sampling rate, and found that more uniform sampling improves performance in low-resource languages, but hurts high-resource languages. 

The reliance on pre-training means that what PLMs encode has far-reaching consequences.
Several works find differences in culture-specific commonsense knowledge in multilingual PLMs, depending on the language used to probe them. These include differences in factual knowledge \cite{kassner-etal-2021-multilingual}, grounding of time-of-day expressions \cite{shwartz2022good}, and social value \cite{lin2021fewshot}. Understanding these differences better will facilitate cultural adaptation and debiasing of NLP systems.

\paragraph{Training.}

The common methodology for training machine learning models (e.g., empirical loss minimisation) relies on maximising average performance across training examples (instead of groups, e.g., languages), which often leads to low minority performance, a phenomenon named
 {\em representation disparity} \cite{hashimoto18a}.
Model performance for minorities is often disregarded in favour of majority groups, as shown for race \cite{Blodgett2017RacialDI}, gender \cite{jorgensen-sogaard-2021-evaluation}, and age \cite{zhang-etal-2021-sociolectal}.
Deriving fair models from biased data is a promising countermeasure \cite{mehrabi2021}. In a cross-cultural setting, methodologies that account for model updates for different groups \cite{hashimoto18a,sagawa-etal-2020-dro,LevyCDS20} could potentially reduce cultural biases.
Most methods to balance for bias require access to protected demographic attributes \cite{sagawa-etal-2020-dro}, and these only partially reflect culture.

In this direction, \citet{zhou2021distributionally} and \citet{ponti-etal-2021-minimax} propose using \emph{Group Distributionally Robust Optimisation} \cite[group DRO;][]{oren-etal-2019-distributionally} to optimise worst-case performance across languages. Similarly, \citet{delhoneux2022worst} used Worst-Case-Aware Curriculum Learning \cite{zhang-etal-2020} to improve group (language) performance parity in cross-lingual dependency parsing.
However, \citet{lent-etal-2021-language} find that group DRO has no benefit in a low-resource settings, namely for Creole languages.
Such techniques can potentially be applied to pre-training multilingual language models with better cross-lingual parity, possibly in addition to improved data sampling.
Furthermore, a fairer model with respect to other attributes, besides the language dimension, can lead to less culturally biased models.
While these measures are widely discussed in a multilingual framework, they are also applicable in a monolingual setting to favour a more equal representation of different cultures in a single language (e.g., a fairer representation of English-speaking communities for English NLP).

\paragraph{Limitations.}

Nonetheless, cross-lingual countermeasures are \emph{culture-agnostic}. In other words, even if we sample languages equally, cross-cultural disparity persists, as data for a given language is also not balanced in terms of sources, and hence cultures. However, one can directly target a more diverse cross-cultural representation by applying the same principles in terms of cultures instead of languages.

Ideally, methods such as data sampling or group DRO should facilitate generalisation of universally-common knowledge from highly-represented cultures, while granting equal representation to minority cultures.
Overemphasising the former is problematic as it contributes to \emph{cultural homogenisation}, while the latter enables mitigation of cross-cultural biases present in the underlying training data and models. There is no substitute for a larger representation of minority cultures in the data used for training and evaluation.

Finally, practical concerns include the level of granularity at which cultural groups are defined and annotated in the data (e.g.,~as metadata attributes or criteria for splitting text corpora). There is a continuum from representing each individual separately (which may raise privacy concerns) to considering large, culturally diverse groups (such as all Spanish speakers) as a homogeneous mass. Again, NLP dataset creators as well as model developers must be aware of the trade-off between generalisation and adaptation here.

\subsection{Translation}\label{sec:translation}

Beyond NLP \textit{for} various cultures, cross-cultural NLP can be used for bridging \textit{between} cultures, investigating cross-cultural communication. The classic example is machine translation (MT) between languages.
In MT, semantic divergences are usually treated as noise, or in any case, as imperfect translations \cite{briakou-carpuat-2021-beyond}. A commonly accepted criterion for translation adequacy is that the semantics of the source are preserved in the output \cite{carl1998meaning,sulem-etal-2015-conceptual}.
However, when applied for translation across cultures, translation may have different objectives. The same meaning may be inappropriate in the target culture, regardless of fluency, and require adjustments: this is referred to as \textit{adaptation} in the translation field \cite{vinay1995comparative}.
\citet{peskov-etal-2021-adapting-entities} observed that translated sentences are often opaque without cultural context (e.g.,  ``I saw Merkel eating a Berliner from Dietsch on the ICE''), and propose adapting entities to the target culture by substituting them with their approximate counterparts. 
As an alternative strategy, adding explanations, for example in the form of appositives \cite{kementchedjhieva-etal-2020-apposcorpus}, can elucidate entities that are known in the source culture but not in the target culture.
These are examples of \textit{cross-cultural translation} \cite{sperber1994cross}.
This kind of adaptation can also be helpful in culturally-situated dialogue, where some users may be less familiar with the common ground of a particular culture; in such cases, \citet{abou2018learning} propose to explain by analogy to concepts from the user's home culture.

Cross-cultural translation is not necessarily always between languages, but could take the form of style transfer within a language \cite{NIPS2017_2d2c8394,prabhumoye-etal-2018-style}.
For example, \citet{jhamtani-etal-2017-shakespearizing} transform text from modern English to Shakespearean English. \citet{roy2015automated} approach personalised marketing by adapting the style of marketing messages for specific audience segments, defined by geographic location and occupation.

Evaluation of cross-cultural translation is challenging, as the task is not always well-defined. In particular, for style transfer, human evaluation is more reliable than automatic evaluation, but still suffers from non-standard evaluation protocols \cite{briakou-etal-2021-evaluating,briakou-etal-2021-review}.
Reference-based automatic evaluation methods are particularly unreliable in this case, as the assumption that there is just one (or a few) \textit{correct} translations is ostensibly violated \citep{Song2013BLEUDD,reiter-2018-structured}. This stresses the need for culture-sensitive human evaluation.

\section{Conclusion}\label{sec:conclusion}

In this paper, we have touched on a number of ways in which cultural knowledge, preferences and values can affect NLP practices.
It is important to acknowledge the breadth of culture in all of these aspects, if we should aspire to a more cross-cultural NLP. However, we do not pretend to cover all relevant aspects of culture in our taxonomy: many works in sociology posture frameworks to better explain and catalogue the specific elements that compose a culture \cite{dant1999material,hofstede2001culture,kendall2005sociology, woodward2007understanding}. We encourage further investigation of their impact on NLP.

Finally, present-day computational science have inherited colonising practices, which in NLP are realised as ``homogenisation of perspectives'' and ``algorithmic monoculture'' \cite{kleinberg2021algorithmic,bommasani2021opportunities}. 
While there is general agreement in the NLP community that we need to represent cultures from outside Western, Educated, Industrialised, Rich, and Democratic (WEIRD) societies \cite{henrich2010weirdest}, 
representation alone is not enough if we do not also allow them to prioritise their goals and values, rather than the goals and values held by the NLP community.
Decolonisation \cite{bird-2020-decolonising,2020,birhane2021towards} aims to dismantle harmful power asymmetries and concepts of knowledge, turning us instead towards a ``pluriversal epistemology of the future'' \cite{Mignolo+2012} that unlike universalisms, acknowledges and supports a wider radius of socio-political, ecological, cultural, and economic needs \cite{2020}.

\section*{Acknowledgements}

The authors thank Desmond Elliott and Vinit Ravishankar for participating in preliminary discussions about the ideas presented in this paper.
We are grateful to Omri Abend, members of the CoAStaL NLP group and the anonymous reviewers for their constructive feedback.
This project has received funding from the European Union's Horizon 2020 research and innovation programme under the Marie Sk\l{}odowska-Curie grant agreement No 801199 (\emph{Heather Lent} and \emph{Emanuele Bugliarello}). This work is also partly funded by the Innovation Fund Denmark (IFD)\footnote{\url{https://innovationsfonden.dk/en}} under File No.\ 0175-00011A (\emph{Ilias Chalkidis} and \emph{Anders Søgaard}). \emph{Miryam de Lhoneux} was funded by the Swedish Research Council (grant 2020-00437).

\section{Ethical Aspects and Broader Impact}\label{sec:ethics}

As discussed in \S\ref{sec:values}, norms and ethics are culture-dependent, and in some cases  there is a conflict between maintaining researchers' and practitioners' ethical values (such as social equity) and multicultural acceptance. It is not our place to settle this conflict or to provide answers, but we stress that asking the question, of which values should prevail in each case, is essential for cross-cultural NLP.
Furthermore, language technology for local communities must involve the members of those communities in an active, participatory manner, in order to decolonise language technology and respect the sovereignty of local people over their data \cite{bird-2020-decolonising,mukhija2021designing}.

\bibliography{anthology,custom}
\bibliographystyle{acl_natbib}

\end{document}